\documentclass[final]{ecai}

\usepackage{mathtools} 

\usepackage{graphicx,verbatim,color,epsfig,graphics,url,multirow}
\usepackage{amsmath,amssymb,amssymb,amsthm,latexsym,thmtools}
\usepackage{bbm}

\usepackage{xspace,setspace}
\usepackage{float}
\usepackage{lineno}
\usepackage{enumitem}
\usepackage{subcaption}
\usepackage[normalem]{ulem}
\usepackage{comment}
\usepackage{adjustbox}         
\usepackage{booktabs,multirow,siunitx}  

\usepackage{todonotes}
\usepackage{algorithm}
\usepackage{algorithmicx}
\usepackage[noend]{algpseudocode}
\usepackage{dashbox}
\usepackage{stmaryrd}
\usepackage{textcomp}
\usepackage{arydshln}
\usepackage{etoolbox}
\usepackage{hyperref}
\usepackage{xfrac}
\usepackage{nicefrac}       
\usepackage{xcolor}         

\usepackage[noabbrev,nameinlink,capitalise]{cleveref}
\usepackage{soul}
\usepackage{pgf}
\usepackage{tikz}
\usepackage{forest}
\usetikzlibrary{arrows,decorations.pathmorphing,decorations.pathreplacing,decorations.footprints,fadings,calc,trees,mindmap,shadows,decorations.text,patterns,positioning,shapes,matrix,fit}
\usetikzlibrary{arrows,shadows,backgrounds}
\usetikzlibrary{arrows.meta}
\usetikzlibrary{positioning,fit}
\usetikzlibrary{automata}
\usetikzlibrary{matrix}
\usetikzlibrary{shapes.symbols,shapes.misc,shapes.arrows}
\usetikzlibrary{matrix,chains,scopes,decorations.pathmorphing}

\newtheoremstyle{nthmstyle}
{3pt}
{3pt}
{}
{}
{\bfseries}
{.}
{.5em}
{}

\theoremstyle{nthmstyle}

\newtheorem{proposition}{Proposition}

\newtheorem{example}{Example}

\crefname{theorem}{Theorem}{Theorems}
\crefname{lemma}{Lemma}{Lemmas}
\crefname{proposition}{Proposition}{Propositions}
\crefname{definition}{Definition}{Definitions}
\crefname{corollary}{Corollary}{Corollaries}
\crefname{example}{Example}{Examples}
\crefname{claim}{Claim}{Claims}
\crefname{assumption}{Assumption}{Assumptions}

\newtheoremstyle{exmpstyle}
{4.0pt} 
{4.0pt} 
{\mdseries} 
{} 
{\itshape} 
{.} 
{ } 
{} 

\theoremstyle{thmstyle}

\crefname{thrm}{Theorem}{Theorems}
\crefname{prop}{Proposition}{Propositions}
\crefname{defn}{Definition}{Definitions}
\crefname{lem}{Lemma}{Lemmas}
\crefname{cor}{Corollary}{Corollaries}
\crefname{crit}{Criterion}{Criteria}
\crefname{clm}{Claim}{Claims}
\crefname{assump}{Assumption}{Assumptions}

\theoremstyle{exmpstyle}

\newtheorem{exmpl}{Example}
\newtheorem*{exmp*}{Example}
{\popQED\endexmpl}
\crefname{rmrk}{Remark}{Remarks}
\crefname{exmp}{Example}{Examples}

\newcommand{\fml}[1]{{\mathcal{#1}}}
\newcommand{\set}[1]{\{ #1 \}}

\newcommand{\tn}[1]{\textnormal{#1}}

\newcommand{\mbf}[1]{\ensuremath\mathbf{#1}}
\newcommand{\msf}[1]{\ensuremath\mathsf{#1}}
\newcommand{\mbb}[1]{\ensuremath\mathbb{#1}}

\newcommand{\Prob}{\ensuremath\tn{Pr}}
\newcommand{\prob}{\Prob}

\providecommand{\Tau}{\mathrm{T}}

\newcommand{\waxp}{\ensuremath\mathsf{WeakAXp}}

\newcommand{\axp}{\ensuremath\mathsf{AXp}}

\newcommand{\ffa}{\ensuremath\mathsf{ffa}}

\newcommand{\delxp}{\ensuremath\mathsf{DelLmPAXp}}

\newcommand{\wpaxp}{\ensuremath\mathsf{WeakPAXp}}
\newcommand{\paxp}{\ensuremath\mathsf{PAXp}}
\newcommand{\mpaxp}{\ensuremath\mathsf{MinPAXp}}
\newcommand{\lmpaxp}{\ensuremath\mathsf{LmPAXp}}

\DeclareMathOperator*{\bigland}{\bigvee}

\DeclareMathOperator*{\lequiv}{\leftrightarrow}
\DeclareMathOperator*{\limply}{\rightarrow}

\definecolor{gray}{rgb}{.4,.4,.4}
\definecolor{midgrey}{rgb}{0.5,0.5,0.5}
\definecolor{middarkgrey}{rgb}{0.35,0.35,0.35}
\definecolor{darkgrey}{rgb}{0.3,0.3,0.3}
\definecolor{darkred}{rgb}{0.7,0.1,0.1}
\definecolor{midblue}{rgb}{0.2,0.2,0.7}
\definecolor{darkblue}{rgb}{0.1,0.1,0.5}
\definecolor{darkgreen}{rgb}{0.1,0.5,0.1}
\definecolor{defseagreen}{cmyk}{0.69,0,0.50,0}

\newcommand{\jnoteF}[1]{}

\newcolumntype{L}[1]{>{\raggedright\let\newline\\\arraybackslash\hspace{0pt}}m{#1}}
\newcolumntype{C}[1]{>{\centering\let\newline\\\arraybackslash\hspace{0pt}}m{#1}}
\newcolumntype{R}[1]{>{\raggedleft\let\newline\\\arraybackslash\hspace{0pt}}m{#1}}

\tikzset{
  0 my edge/.style={densely dashed, my edge},
  my edge/.style={-{Stealth[]}},
}

\setlist{nosep}

\providetoggle{long}
\settoggle{long}{false}

\newcommand*\patchAmsMathEnvironmentForLineno[1]{%
  \expandafter\let\csname old#1\expandafter\endcsname\csname #1\endcsname
  \expandafter\let\csname oldend#1\expandafter\endcsname\csname end#1\endcsname
  \renewenvironment{#1}%
     {\linenomath\csname old#1\endcsname}%
     {\csname oldend#1\endcsname\endlinenomath}}%
\newcommand*\patchBothAmsMathEnvironmentsForLineno[1]{%
  \patchAmsMathEnvironmentForLineno{#1}%
  \patchAmsMathEnvironmentForLineno{#1*}}%
\AtBeginDocument{%
\patchBothAmsMathEnvironmentsForLineno{equation}%
\patchBothAmsMathEnvironmentsForLineno{align}%
\patchBothAmsMathEnvironmentsForLineno{flalign}%
\patchBothAmsMathEnvironmentsForLineno{alignat}%
\patchBothAmsMathEnvironmentsForLineno{gather}%
\patchBothAmsMathEnvironmentsForLineno{multline}%
}

\begin{document}

\begin{frontmatter}

\title{Locally-Minimal Probabilistic Explanations}


\author[A]{\fnms{Yacine}~\snm{Izza}\thanks{Corresponding Author. Email: izza@comp.nus.edu.sg}}
\author[B]{\fnms{Kuldeep S.}~\snm{Meel}}
\author[C]{\fnms{Joao}~\snm{Marques-Silva}} 

\address[A]{CREATE, NUS, Singapore}
\address[B]{University of Toronto, Canada}
\address[C]{ICREA, University of Lleida, Spain}

%
%
%
\begin{abstract}
  Explainable Artificial Intelligence (XAI) is widely regarding as a
  cornerstone of trustworthy AI. Unfortunately, most work on XAI
  offers no guarantees of rigor. In high-stakes domains, e.g.\ uses of
  AI that impact humans, the lack of rigor of explanations can have
  disastrous consequences. 
  Formal abductive explanations offer crucial guarantees of rigor and
  so are of interest in high-stakes uses of machine learning (ML).
  One drawback of abductive explanations is explanation size,
  justified by the cognitive limits of human decision-makers.
  Probabilistic abductive explanations (PAXps) address this
  limitation, but their theoretical and practical complexity makes
  their exact computation most often unrealistic.
  This paper proposes novel efficient algorithms for the computation
  of locally-minimal PXAps, which offer high-quality approximations of
  PXAps in practice.
  The experimental results demonstrate the practical efficiency of the
  proposed algorithms.
\end{abstract}

\end{frontmatter}

\section{Introduction} \label{sec:intro}

For more than a decade, progress in Machine Learning (ML) has been
revolutionary. However, such progress is most often achieved through
the use of highly complex ML models, which human decision-makers are
unable to comprehend.
In recent years, eXplainable Artificial Intelligence (XAI) has been
proposed as a solution to deliver trustworthy AI, thus enabling the
use of complex ML models in high-stakes domains.
Unfortunately, most work on XAI offers little to no guarantees of
rigor. Unsurprisingly, there exists growing evidence to the
limitations and perils of deploying XAI solutions that offer no
guarantees of rigor~\cite{ms-iceccs23}.

Recent work has advocated the use of formal, logic-based
explanations~\cite{msi-aaai22,darwiche-lics23}. Formal XAI offers the
strongest guarantees of rigor, but it also exhibits important
limitations. One of the best-known limitations is scalability, due to
the complexity of logic-based reasoning. While some progress has been
observed for tree
ensembles~\cite{ims-ijcai21,Audemard-aaai22,IgnatievIS022,AudemardLMS23},
neural networks have represented a significant
challenge for formal XAI~\cite{inms-aaai19}. Nevertheless, recent
work~\cite{barrett-nips23} showed important progress towards achieving
scalability of formal XAI in highly complex ML models.
Another key limitation of formal XAI is the size of explanations,
which can often be well beyond the cognitive limits of human decision
makers~\cite{miller-pr56}.
Probabilistic explanations~\cite{kutyniok-jair21} aim at finding
explanations of smaller size, and so represent one proposed solution
for the size limitation of formal explanations.
%
%
%
Probabilistic abductive explanations trade off the strong theoretical
guarantees of rigor of abductive explanations for smaller explanation
sizes, while still ensuring the quality of approximate explanations.
Unfortunately, the complexity of finding probabilistic abductive
explanations is in general unwiedly~\cite{kutyniok-jair21}, e.g.\ a
practical algorithm should be expected to call a counting oracle
exponentially many times. Even for restricted families of classifiers,
e.g.\ decision trees, it is known that the computation of
probabilistic abductive explanations is computationally hard, albeit
solvable with a limited number of calls to an
NP-oracle~\cite{barcelo-nips22,ihincms-ijar23}. Clearly, the case of
decision trees, while still computationally hard, is drastically
easier to solve in practice.

One key difficulty in computing probabilistic abductive explanations 
results from the non-monotonicity of the sets representing or not a 
probabilistic explanation. This lack of monotonicity requires
the analysis of all possible subsets of features in order to find a
subset-minimal set. In such a situation, one approximation to
subset-minimality is local minimality, i.e. sets that are minimal if
at most one feature is allowed to be removed. A surprising
experimental observation~\cite{ihincms-ijar23} is that, in the case of
decision trees and other graph-based classifiers, locally-minimal
explanations are in most cases also subset-minimal explanations --- reported 
results show for decision trees that in 99.8\% cases computed approximate 
explanations are proved that are subset minimal. 

%

Given these earlier experimental observations, one solution towards
devising practical algorithms for approximating probabilistic
abductive explanations is to compute locally-minimal explanations,
thereby eliminating the need to analyze all possible subsets of a
target set of features.
This paper proposes two new algorithms for computing approximate
locally-minimal explanations, one based on approximate model counting,
and the other based on sampling with probabilistic guarantees.
The experimental results support the practical efficiency of the
proposed algorithms for two case studies of classifiers random forest (RF)
and binarized neural network (BNN) --- results on BNNs show that 
explanation length of our solution drops by one third to two third  of 
 abductive explanation size.

The paper is organized as follows.
\cref{sec:prelim} introduces the notation and definitions used in the
paper.
\cref{sec:enc} overviews the logic encodings of classifiers used in
later sections.
\cref{sec:approach} describes the approach proposed in this paper.
The experimental results are discussed in~\cref{sec:res}.
Finally, the paper concludes in~\cref{sec:conc}.

\section{Background} \label{sec:prelim}
Here we introduce the notation and background on formal 
XAI and the case study of two classifier families,  which 
enable (propositional) logical encoding.  

\subsection{Classification problems}
This paper considers classification problems, which are defined on a
set of features (or attributes) $\fml{F}=\{1,\ldots,m\}$ and a set of
classes $\fml{K}=\{c_1,c_2,\ldots,c_K\}$.
Each feature $i\in\fml{F}$ takes values from a domain $\mbb{D}_i$.
In general, domains can be categorical or ordinal, with values that
can be boolean, integer or real-valued.
Feature space is defined as
$\mbb{F}=\mbb{D}_1\times{\mbb{D}_2}\times\ldots\times{\mbb{D}_m}$;
$|\mbb{F}|$ represents the total number of points in $\mbb{F}$.
For boolean domains, $\mbb{D}_i=\{0,1\}=\mbb{B}$, $i=1,\ldots,m$, and
$\mbb{F}=\mbb{B}^{m}$.
The notation $\mbf{x}=(x_1,\ldots,x_m)$ denotes an arbitrary point in
feature space, where each $x_i$ is a variable taking values from
$\mbb{D}_i$. The set of variables associated with features is
$X=\{x_1,\ldots,x_m\}$.
Moreover, the notation $\mbf{v}=(v_1,\ldots,v_m)$ represents a
specific point in feature space, where each $v_i$ is a constant
representing one concrete value from $\mbb{D}_i$. 

An ML classifier $\mbb{M}$ is characterized by a (non-constant)
\emph{classification function} $\kappa$ that maps feature space
$\mbb{F}$ into the set of classes $\fml{K}$,
i.e.\ $\kappa:\mbb{F}\to\fml{K}$.
An \emph{instance} 
denotes a pair $(\mbf{v}, c)$, where $\mbf{v}\in\mbb{F}$ and
$c\in\fml{K}$, with $c=\kappa(\mbf{v})$. 
Given the above, we represent a classifier $\fml{M}$ by a tuple
$\fml{M}=(\fml{F},\mbb{F},\fml{K},\kappa)$.

Moreover, and given a concrete instance $(\mbf{v},c)$, an explanation
problem is represented by a tuple $\fml{E}=(\fml{M},(\mbf{v},c))$.

\subsection{Binarized Neural Network (BNNs)}
Binarized Neural Networks (BNNs)~\cite{bengio-nips16,bengio-corr16}
are a widely used family of neural networks. BNNs exhibit a number of
important features that allow them to be deployed into embedded
devices~\cite{McDanelTK17,KungZWCM18}. 
Concretely, a BNN is composed of a number of layers of neurons. The neurons 
of the intermediate layers (hidden layers) compute a mapping function: 
$\set{-1, 1}^{n}\rightarrow\set{-1,1}^{m}$  on input  $x_i\in\set{-1, 1}^{n}$, whilst 
the neurons of the last layer (output layer) map a binary tensor to real domain: 
$\set{-1, 1}^{n_d}\rightarrow\mbb{R}^K$ on input $x_i\in\set{-1, 1}^{n_d}$.
Furthermore, outputs of intermediate layer are obtained with applying 
three transformations on the input tensor $x$: a linear transformation (LIN), 
batch normalization (BatchNorm) and binarization (BIN), such that 
$  x^{i+1} = \textsc{Bin} ( \textsc{BatchNorm} (\textsc{Lin}(x))) $, where:
\textsc{Lin}($x^i$) = $W^i x^i + b^i$, where $W^i \in \set{-1, 1}^{m \times n}$ and 
$b^i \in \mbb{R}^n$, 
\textsc{BatchNorm}($y_j$) =  $\alpha^i_j \left( \frac{y_j - \mu_j^i}{\sigma^i_j} \right)$, where  
      				$\alpha^i_j,   \mu_j^i, \sigma^i_j, y_j \in \mbb{R}^m$, 
and 	 \textsc{Bin}($z$) = $ sign(z) \in \set{-1,1}^m$.	
Lastly, the output layer applies a linear transformation before 
\emph{argmax} mapping that picks a class to predict. 
(Note that~\cite{narodytska-iclr20,rinard-nips20} propose to use ternary 
neural network in order to generate sparse matrices, hence  we have 
$W^i \in \set{-1, 0, 1}$.

\subsection{Random Forests (RFs)}
Random Forests
(RFs)~\cite{breiman2001random,YangWJZ20,Zhang19,GaoZ20,FengZ18a,ZhouF17}
are an example of tree ensemble ML models, which find a wide range of
practical applications.
Although RFs are known to lack in interpretability, recent
work~\cite{ims-ijcai21,iisms-aaai22} demonstrates  that large tree
ensembles can be efficiently analyzed with logical-based
reasoners like SAT/SMT oracles, and so can be explained.
Conceptually, an RF is collection of decision trees (DTs), where each tree 
$\fml{T}_i$, $i \in \{1, \dots, T\}$ of the ensemble $\fml{T}$ is trained on a randomly 
selected subset of the training data so that the trees of the RF are
not correlated. 
(In contrast to a single DT, RFs are less prone to over-fitting and so
offer in general better accuracy on test data~\cite{breiman2001random}.)
Similar to the original proposal for RFs~\cite{breiman2001random}, the
predictions of a RF classifier are made by majority vote of trees, 
that is each tree predicts for a class and the class with largest score
is picked. Other solutions could be considered, e.g. weighted voting.

\subsection{Logic-based Explainability}
Formal explainability aims to answer the question ``{\it why}'' 
a prediction is made by a classifier by identifying a selection 
of the input features that are responsible of the prediction, and 
consequently can be referred as feature selection explanations. 
More recently, formal {\it feature attribution} explanations 
\cite{Ignatiev-CoRR23a,Izza-aaai24a,Ignatiev-CoRR23b} have been 
introduced, which provides the importance score of each 
feature included the explanation set.
We overview the formal definition of both type of explanations.   
  

\paragraph{Abductive explanations.}
Prime implicant (PI) explanations~\cite{darwiche-ijcai18} denote a
minimal set of literals (relating a feature value $x_i$ and a constant
$v_i\in\mbb{D}_i$) 
that are sufficient for the prediction. PI-explanations are related
with abduction, and so are also referred to as (feature selection) abductive 
explanations (AXp's)~\cite{inms-aaai19}.
Formally, given $\mbf{v}=(v_1,\ldots,v_m)\in\mbb{F}$ with
$\kappa(\mbf{v})=c$,
a set of features $\fml{X}\subseteq\fml{F}$ is a \emph{weak abductive
  explanation}~\cite{cms-cp21} (or weak AXp) if the following
predicate holds true:
\begin{align} \label{eq:waxp}
    \waxp(\fml{X}) 
    \:{:=} ~ & 
    \forall(\mbf{x}\in\mbb{F}).  \nonumber \\
    &
    \left[
      \bigwedge\nolimits_{i\in{\fml{X}}}(x_i=v_i)
      \right]
    \limply(\kappa(\mbf{x})=c)
\end{align}
The classifier assigns the same class to all feature-vectors which agree with $\mbf{v}$ 
on features in a weak AXp $\fml{X}$. Such sets $\fml{X}$ may contain features which do not 
contribute to the decision, which leads to the following definition.
A set of features $\fml{X}\subseteq\fml{F}$ is an
\emph{abductive explanation} (or (plain) AXp) if the following
predicate holds true: 
\begin{align} \label{eq:axp}
  \axp(\fml{X})\quad{:=}\quad&
  \waxp(\fml{X}) ~\land \nonumber \\
  &\forall(\fml{X}'\subsetneq\fml{X}).
  \neg\waxp(\fml{X}')
\end{align}
Clearly, an AXp is any weak AXp that is subset-minimal (or
irreducible).
It is straightforward to observe that the definition of
predicate $\waxp$ is monotone, and so an AXp can instead be defined as
follows:
\begin{align} \label{eq:axp2}
  \axp(\fml{X}) \quad{:=}\quad&
  \waxp(\fml{X}) ~\land \nonumber \\
    &\forall(j\in\fml{X}).
    \neg\waxp(\fml{X}\setminus\{j\})
\end{align}
(Throughout the document, we will drop the parameterization associated
with each predicate, and so we will write $\axp(\fml{X})$ instead of
$\axp(\fml{X})$, when the parameters are
clear from the context.)

Observe that an instance $(\mbf{v}, c)$ can admit more that one 
AXp (which are worst-case exponential). 
We denote by $\mbb{A}(\mbf{v},c)$ the set of all possible AXp's of 
$(\mbf{v}, c)$.   

Logic-based explainability is covered in a number of recent works.
Explaining tree ensembles is 
studied in~\cite{ims-ijcai21,Audemard-aaai22,AudemardBLM23,AudemardLMS23}. 
Probabilistic explanations are investigated
in~\cite{kutyniok-jair21,barcelo-nips22,ihincms-ijar23}.
%
%

\paragraph{Running Example.}
We consider a decision tree (DT) as a running example to illustrate 
the different class of explanations (AXp's and Probabilistic AXp's).
\begin{figure}[ht]
   \centering
    \tikzset{every label/.style={xshift=-0.35ex,
  yshift=-6.225ex,
  text width=1ex,
  align=right, inner sep=1pt, font=\tiny, text=midblue}}
\tikzset{tlabel/.style={xshift=0.25ex, yshift=2ex, text width=1ex,
    align=right, inner sep=1pt, font=\tiny, text=midblue}}
\forestset{
  BDT/.style={
    for tree={
      l=1.35cm,s sep=1.5cm,
      if n children=0{}{circle},
      draw,
      edge={
        my edge
      },
      if n=1{
        edge+={0 my edge},
      }{},
    }
  },
}
\begin{forest}
  BDT
  [$x_{1}$, label={1}
    [$x_{2}$, s sep=0.75cm, label={2},
      edge label={node[near start,left,xshift=-0.75pt] {{\scriptsize$\in\{1\}$}}}
      [{\footnotesize\color{darkred}$\ominus$},
        label={[xshift=0.25ex,yshift=1.875ex]4},
        edge label={node[near start,left,xshift=-0.5pt]
          {{\scriptsize$\in\{1,2\}$}}}]
      [{\footnotesize\color{darkgreen}$\oplus$},
        label={[xshift=0.25ex,yshift=1.875ex]5},
        edge label={node[near start,right,xshift=-1pt]
          {{\scriptsize$\in\{3..5\}$}}}]
    ]
    [$x_2$, s sep=0.75cm, label={3},
      edge label={node[near start,right,xshift=1pt]
        {{\scriptsize$\in\{2..4\}$}}}
      [{\footnotesize\color{darkgreen}$\oplus$},
        label={[xshift=0.25ex,yshift=1.875ex]6},
        edge label={node[near start,left,xshift=-1pt]
          {{\scriptsize$\in\{1\}$}}}]
      [$x_3$, label={7},
        edge label={node[near start,right,xshift=-1pt]
          {{\scriptsize$\in\{2..5\}$}}}
        [{\footnotesize\color{darkred}$\ominus$},
          label={[xshift=0.25ex,yshift=1.875ex]8},
          edge label={node[pos=0.6,left,xshift=-1pt]
            {{\scriptsize$\in\{1..3\}$}}}]
        [{\footnotesize\color{darkgreen}$\oplus$},
          label={[xshift=0.25ex,yshift=1.875ex]9},
          edge label={node[pos=0.6,right,xshift=0.25pt]
            {{\scriptsize$\in\{4\}$}}}]
      ]
    ]
  ]
\end{forest}
  \caption{%
  	Example decision tree. Nodes of the tree are enumerate 
  	from 1 to 9; a tree path is represented by a sequence of nodes  
	(e.g.\ $\langle1,2,4\rangle$, $\langle1,3,6\rangle$, etc).
  } 
  \label{fig:runex01:dt}
  \bigskip
\end{figure}
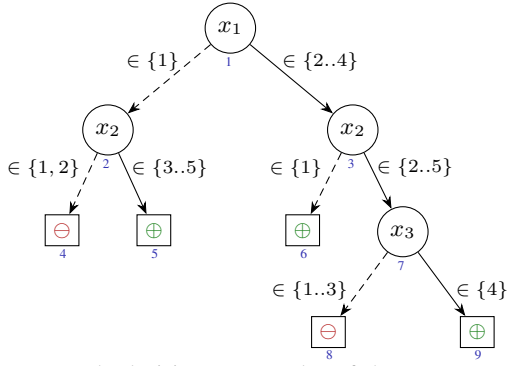

\begin{example} \label{ex:runex01}
    \cref{fig:runex01:dt} shows an example decision tree, adapted from 
    a toy example shown in~\cite{ihincms-ijar23}, that admits 
    a vector input of three features $\set{x_1, x_2, x_3}$ where the feature domains 
    are, respectively, $\mbb{D}_1 =  \mbb{D}_3 = \set{1, \dots,4}$ and 
    $\mbb{D}_2 = \set{1, \dots,5}$,  and predicts class $\oplus$ or $\ominus$. 
    %
    %
    Consider instance  $\mbf{v} = (2, 3, 1)$, with $c = \kappa(\mbf{v}) = \ominus$. 
    (The point $\mbf{v}$ is consistent with the path of nodes $\langle1,3,7,8\rangle$).  
    Clearly, one can observe that $\fml{X} = \set{1, 2, 3}$ is an AXp for $\mbf{v}$. i.e.\ 
    observe that if we drop $x_1$, then $(x_2 = 3~\wedge~x_3 = 1)$ is consistent 
    with the path $\langle1,2,5\rangle$, or $x_2$ then  $(x_1 = 2~\wedge~x_3 = 1)$ 
    is consistent with  $\langle1,3,6\rangle$, or $x_3$ then $(x_1 = 2~\wedge~x_2 = 3)$ 
    is consistent with  $\langle1,3,7,9\rangle$.	
\end{example}

\paragraph{Formal Feature Attribution AXp's.} 
Formal feature attribution (FFA)~\cite{Ignatiev-CoRR23a,Izza-aaai24a} 
abductive explanations are  an extension of (plain) abductive 
explanations, in sense that they provide additional information about 
the importance score of the features in the AXp's. 
Feature atribution AXp can be seen as an aggregation of all possible (or partial collection in 
the context of approximate FFA) AXp's~\cite{Izza-aaai24a} for a given instance 
$\mbf{v}$ to be explanations.
Concretely, FFA of a feature $i\in\fml{F}$, denoted $\ffa(i)$ is defined as 
the proportion of AXp's  where it occurs, i.e.\   

\begin{equation*}
 \ffa(i) = \frac{|\{ \fml{X} \mid \fml{X}\in\mbb{A}(\mbf{v}, c) , i\in\fml{X} \}|}{\mbb{A}(\mbf{v}, c)}  
\end{equation*}

We can define now a formal feature attribution abductive explanation (FFAXp) 
as a set of features $\fml{X}\subseteq\fml{F}$ for which $i\in\fml{X}, \ffa(i) > 0$ 
and $i\in\fml{F}\setminus{X}, \ffa(i) = 0$.
Clearly, an FFAXp $\fml{X}$ is also a weak AXp but not necessarily an AXp. 
Note that in practice FFAXp's are (much) larger than AXp's, hence  
our interest to seek for more succinct approximation of FFAXp's.

\subsection{Formal Probabilistic Explanations}
We follow the recent accounts on formal probabilistic 
explanations~\cite{ihincms-ijar23}.   

\begin{table*}[t]
  \centering
  \renewcommand{\arraystretch}{1.05}
  \renewcommand{\tabcolsep}{0.325em}
    \begin{tabular}{cccccccc} \toprule
      $\fml{S}$ & 
      $\waxp$? & $\axp$? &
      $\prob_{\mbf{x}}(\kappa(\mbf{x})=\ominus|(\mbf{x}_{\fml{S}}=\mbf{v}_{\fml{S}}))$ &
      $\wpaxp$? & $\paxp$? & $\lmpaxp$? & $\mpaxp$?
      \\ \toprule
      $\{1,2,3\}$ & 
      Yes & Yes & 
      $1\ge\Tau$ &
      Yes & No & No & No
      \\ \cmidrule(lr){1-1} \cmidrule(lr){2-3} \cmidrule(lr){4-4} \cmidrule(lr){5-8}
      $\{1,2\}$ & 
      No & No &
      $\sfrac{3}{4}=0.75\ge\Tau$ &
      Yes & Yes & Yes & No
      \\ \cmidrule(lr){1-1} \cmidrule(lr){2-3} \cmidrule(lr){4-4} \cmidrule(lr){5-8}      
      $\{1,3\}$ & 
      No & No &
      $\sfrac{4}{5}=0.8\ge\Tau$ &
      Yes & No & No & No
      \\ \cmidrule(lr){1-1} \cmidrule(lr){2-3} \cmidrule(lr){4-4} \cmidrule(lr){5-8}          
      $\{2,3\}$ & 
      No & -- &
      $\sfrac{3}{4}=0.75\ge\Tau$ &
      Yes & No & No & No
      \\ \cmidrule(lr){1-1} \cmidrule(lr){2-3} \cmidrule(lr){4-4} \cmidrule(lr){5-8}
      $\{1\}$ & 
      No & -- &
      $\sfrac{12}{20}=0.6\le\Tau$ &
      No & No & No & No
      \\ \cmidrule(lr){1-1} \cmidrule(lr){2-3} \cmidrule(lr){4-4} \cmidrule(lr){5-8}       
      $\{2\}$ & 
      No & -- &
      $\sfrac{9}{16}=0.5625\le\Tau$ &
      No & No & No & No
      \\ \cmidrule(lr){1-1} \cmidrule(lr){2-3} \cmidrule(lr){4-4} \cmidrule(lr){5-8}
      $\{3\}$ & 
      No & -- &
      $\sfrac{14}{20}=0.7\ge\Tau$ &
      Yes & Yes & Yes & Yes
      \\ 
      \bottomrule
    \end{tabular}
  \caption{Examples of sets of fixed features given instance $\mbf{v}=(2,3,1)$
  (s.t. $\kappa(\mbf{v})=\ominus$)  and $\Tau=0.7$.}
  \label{tab:cprob}
  \bigskip
\end{table*}

\paragraph{Probabilistic AXp.} 
A \emph{weak probabilistic} AXp (or weak PAXp)~\cite{iincms-corr22,ihincms-ijar23} 
for a given instance 
$\mbf{v}$ with $\kappa(\mbf{v}) = c$, is a
a set of features fixed to values of  $\mbf{v}$ for which the conditional 
probability of predicting the correct class $c$ is no less than 
a given threshold $\Tau \in [0, 1]$.
Thus, $\fml{X}\subseteq\fml{F}$ is a weak PAXp if the following
predicate holds true,
\begin{align} \label{eq:wpaxp}
  \wpaxp&(\fml{X})  
  \nonumber \\
  :=\,\: & \prob_{\mbf{x}}(\kappa(\mbf{x})=c\,|\,\mbf{x}_{\fml{X}}=\mbf{v}_{\fml{X}})
  \ge \Tau
  \\[1.0pt]
  :=\,\: &\frac{%
    |\{\mbf{x}\in\mbb{F}:\kappa(\mbf{x})=c\land(\mbf{x}_{\fml{X}}=\mbf{v}_{\fml{X}})\}|
  }{%
    |\{\mbf{x}\in\mbb{F}:(\mbf{x}_{\fml{X}}=\mbf{v}_{\fml{X}})\}|
  }
  \ge\Tau \nonumber
\end{align}
which means that the fraction of the number of points predicting the
target class and consistent with the fixed features (represented by
$\fml{X}$), given the total number of points in feature space
consistent with the fixed features, must exceed $\Tau$.
(Note that  $\prob_{\mbf{x}}(\kappa(\mbf{x})=c\,|\,\mbf{x}_{\fml{X}}=\mbf{v}_{\fml{X}})$ 
is commonly referred as the probability, precision or accuracy of  $\fml{X}$ 
\cite{guestrin-aaai18,vandenbroeck-ijcai21,kutyniok-jair21}, thus 
we will use these terms interchangeably throughout the paper.
For notation simplicity, we will also denote by $\prob_{\mbf{x}}(\fml{X})$ 
the probability/precision  of an explanation $\fml{X}$ for $\fml{E}$.) 
Moreover, a set $\fml{X}\subseteq\fml{F}$ is a \emph{probabilistic}
AXp (or (plain) PAXp) if the following predicate holds true,
\begin{align} \label{eq:paxp}
  \paxp&(\fml{X}) \::= \nonumber \\
  &\wpaxp(\fml{X}) \:\:\land \\
  &\forall(\fml{X}'\subsetneq\fml{X}). %
  \neg\wpaxp(\fml{X}') \nonumber
\end{align}
Thus, $\fml{X}\subseteq\fml{F}$ is a PAXp if it is a weak PAXp that is
also subset-minimal, 

As can be observed, the definition of weak PAXp (see~Eq.~\eqref{eq:wpaxp})
does not guarantee monotonicity. In turn, this makes the computation
of (subset-minimal) PAXp's harder.
With the purpose of identifiying classes of weak PAXp's that are
easier to compute, it will be convenient to study
\emph{locally-minimal} PAXp's.

\paragraph{Locally-minimal Probabilistic AXp.} 
A set of features
$\fml{X}\subseteq\fml{F}$ is a locally-minimal PAXp if,
\begin{align} \label{eq:lmpaxp}
  \lmpaxp&(\fml{X}) \::= \nonumber \\
  &\wpaxp(\fml{X}) \:\:\land \\
  &\forall(j\in\fml{X}). %
  \neg\wpaxp(\fml{X}\setminus\{j\}) \nonumber
\end{align}
As observed earlier, because the predicate $\waxp$
is monotone, subset-minimal AXp's match locally-minimal AXp's. An
important practical consequence is that most algorithms for computing
one subset-minimal AXp, will instead compute a locally-minimal AXp,
since these will be the same.
Nevertheless, a critical observation is that in the case of
probabilistic AXp's (see~Eq.~\eqref{eq:wpaxp}), the predicate $\wpaxp$ is
\emph{not} monotone.
Thus, there can exist locally-minimal PAXp's that are not
subset-minimal PAXp's. 

An easy observation is that LmPAXp definition also applies for FFAXp's. 
We denote  by LmPFFAXp's the {\it locally-minimal probabilistic} FFAXp's 
to distinguish between probabilistic AXp's and FFAXp's.

Finally, the predicate $\mpaxp$ defines the {\it minimum-size} PAXp’s 
(or a smallest PAXp’s). Note that in practice, explanation size 
of PAXp's and LmPAXp's are often not much larger than  
minimum-size PAXp’s. This observation is supported by the empirical results 
on different families of classifiers shown in~\cite{ihincms-ijar23}.  


\begin{example}
   Let consider again DT and instance $\mbf{v}$ of \cref{ex:runex01}. 
  \cref{tab:cprob} summarizes the computation of 
   $\prob_{\mbf{x}}(\kappa(\mbf{x})=\ominus|(\mbf{x}_{\fml{S}}=\mbf{v}_{\fml{S}}))$
    for different  sets of features  $\{1,2,3\}$, $\{1,3\}$, $\{2\}$ and $\{3\}$. Moreover,  it reports 
    whether the analyzed subset $\fml{S}$ is a (weak) AXp, 
  (Weak) PAXp, a MinPAXp or an LmPAXp. 
  (The analysis of all points in feature space is omitted for brevity.) 
  The set $\set{1, 2}$ is an LmPAXp and a PAXp (and necessarily a weak PAXp) 
  since the he probability  of predicting $\ominus$ is greater than $\Tau$ and its 
  subsets $\set{1}$ and $\set{2}$ are not weak PAXp.
  The set $\{1,3\}$ and $\{2,3\}$  are weak PAXp but not PAXp or LmPAXp, 
  since that their respective probabilities are are greater than $\Tau$ but 
  the subset $\set{3}$ is also a weak PAXp. Moreover,  $\set{3}$ is a minimum PAXp 
  (and necessarily a PAXp). 
  Observe that the order to analyzing features when extracting an  LmPAXp 
  is important in order to find the smallest explanation. Assume for instance  
  we traverse the features in the order $\langle3, 2, 1\rangle$.  Then, 
  we can easily see that the final LmPAXp would 
  be $\fml{X} = \set{1, 2}$ of size 2 and not $\fml{X} = \set{3}$ of size 1.  
\end{example}

\section{Logic Encodings} \label{sec:enc}
In this section
we detail the encodings for both random forests (with majority voting)
and binarized neural networks.
It should be noted that the proposed declarative encodings on both
random forests and binarized neural networks are polynomial on the
size of the original ML classifier.

\subsection{BNN Encoding}
%
Recent work 
\cite{narodytska-aaai18,ChengNHR18,narodytska-iclr20,rinard-nips20} 
on formal verification of BNNs have 
proposed to formulate a number of verification queries on BNNs into 
propositional formulas or  pseudo-Boolean (PB) formulas.  
We extend the SAT/PB -based encoding for robustness checking for 
computing explanation in BNNs.
We underline that PB constraints can be efficiently converted 
into CNF formulas. Therefore, this work focuses on PB encodings, which 
can be translated into SAT, instead of an immediate SAT encoding. 
Roughly, encoding a BNN is performed by encoding each neuron 
in the network with a constraint of the form:
\begin{equation} \label{eq:neuron}
    y \lequiv   \sum\nolimits^{n}_{i=1} w_i \cdot l_i  \ge b,  ~~~ y\in\set{0,1} 
\end{equation}	
where $l_i$ are Boolean variables, $w_i$ and $b$ are integer 
variables\footnote{%
For more details on how the inequality is calculated (coefficients, bound, etc) 
the reader can refer to~\cite{narodytska-aaai18} (Sec. 4) 
or \cite{narodytska-iclr20} (Sec. 3.1).} and  
$(2\cdot y - 1)$ is the output of the (internal) neuron. 
Note that~\cref{eq:neuron} can be easily transformed into 
a \emph{reified cardinality constraint} with unary coefficients, and 
thus be encoded  with sequential encoders~\cite{Sinz05a,Sinz05b} in 
clausal form, which yields to a CNF representation of the BNN.  

Besides, \cref{eq:neuron} can be rewritten into two PB constraints:
%
%
\[
\begin{array}{lll}
    \sum\nolimits^{n}_{i=1} w_i l_i + (b + N) \cdot \neg y \ge b &
    \text{for} & 
    y  \limply \sum\nolimits^{n}_{i=1} w_i l_i  \ge b   \\ [3pt]
   \sum\nolimits^{n}_{i=1}  w_i l_i + (b - N) \cdot  y  \le b & 
   \text{for}   & 
   \neg y  \limply \sum\nolimits^{n}_{i=1} w_i l_i  < b  \\
\end{array}
\]
with $N = \sum\nolimits^{n}_{i=1} |w_i|$.

Next, we need to encode 
the output layer of the neural network, i.e. $\kappa(\mbf{x})$.
Let $s_j$ Boolean variables representing class $c_j\in\fml{K}$, 
$W^i_j$ is a tensor of weights and $b_j$ is a bias of a neuron $j$ 
in the last layer of the BNN. Hence,  
a target class $c_j$ is picked ($s_j = 1$) iff,
\begin{align}
   \bigwedge\nolimits_{k\neq j}  y_k \lequiv \left( (W^i_k - W^i_j)\cdot l_i \ge ( b_k - b_j ) \right) 
   \nonumber \\ 
   \wedge  \left( s_j \lequiv \bigwedge\nolimits_{k\neq j}  \neg y_k \right)
\end{align}

\subsection{RF Encoding}
%
We adapt the propositional encoding of RFs 
proposed in~\cite{ims-ijcai21} to encode $\kappa(\mbf{x})$  
entails the prediction of input $\mbf{v}$, 
in order to be able to compute the number of models 
satisfying this entailment. 
Concretely, we need to encode the RF structure   (i.e. trees and 
feature-domain encodings) and the prediction function $\kappa(\mbf{x})$ 
(i.e. majority vote of the ensemble model) into a SAT problem.
As results, each tree $\fml{T}_i$ of the model is encoded 
as a set of implications expressing the paths $\fml{P}_i$ of $\fml{T}_i$, 
as follow:  
$\bigland\nolimits_{P \in \fml{P}_i} \left( \bigland\nolimits_{l \in \Lambda(P)} l \lequiv p_{ij} \right)$, 
where   $\Lambda(P)$ is the set of literals of path $P$ and 
$p_{ij}$ represents the class label $c_j$ of $P$, i.e. 
$p_{ij} = 1$ if the class $c_j$ is predicted when $P$ is consistent.
Next,  we outline the encoding of majority vote for $\kappa(\mbf{x})$. 
%
%
Then,  the idea is to enforce the number of votes 
for the target class $c_j$ to be the largest one for $\kappa(\mbf{x})$:
  \begin{equation}\label{eq:RF}
      \bigwedge\nolimits_{k \neq j}       
      \sum\nolimits_{i=1}^{M}p_{ij}+\sum\nolimits_{i=1}^{M}\neg{p_{ik}} + b_k\ge{M+1} 
  \end{equation}
    Remark that $b_k$ is fixed $b_k = 1$ iff $c_k \prec c_j$. Thus, for the case 
    where $c_k \succ c_j$ the cardinality constraint enforces the inequality
    $\left( \sum\nolimits_{i=1}^{M}p_{ij} > \sum\nolimits_{i=1}^{M}\neg{p_{ik}} \right)$, 
    then RHS equals to $M+1$.


%
\section{Approximate Locally Minimal Probabilistic Explanations} \label{sec:approach}
We are interested in (approximately) computing LmPAXp's for 
complex ML models, including tree ensembles  and binarized 
neural networks. Therefore, the difficulty is to compute 
or estimate 
the probability of the logical formula representing the model 
to be satisfiable.
Clearly, applying exact model counting for large propositional 
formulas  will be infeasible in practice. We solve this problem 
using either approximate model counting or sampling, with strongly
probably approximately correct (PAC) guarantees.
A procedure implementing one of the approaches will serve 
to check whether a candidate set of features $\fml{X}$ is 
a weak PAXp.
Additionally, two linear search techniques are outlined in this 
section for extracting LmPAXp's by instrumenting the model counting 
or sampling subroutine.

\subsection{Approximate Model Counting}
%
The approximate model counter serves as oracle in the linear 
search analysis to compute the number of points $\mbf{x}\in\mbb{F}$,
such that the condition $\kappa(\mbf{x}) = \kappa(\mbf{v})$ and 
$\mbf{x}_\fml{X} = \mbf{v}_\fml{X}$ holds.
We formulate this as SAT (or pseudo-Boolean) formula 
(as described earlier in~\cref{sec:enc}) 
and instrument an oracle call to a ({\it probably approximately correct} (or PAC)) 
counter ApproxMC\footnote{
ApproxMC is a state-of-the-art approximate model counter, that 
scales for large problem instances and provides rigorous 
approximation guarantees.} 
\cite{ChakrabortyMV16,SoosM19,SoosGM20,MeelA20} 
(resp.\ ApproxMCPB~\cite{YangM21,YangM23} for PB formulas) 
that returns an ($\epsilon, \delta$)-approximate  number of solutions 
$\eta^\ast$ of the input formula, such that 
\[ \prob \left[ \frac{\eta}{1+\epsilon} \le \eta^\ast \le  (1+\epsilon)\times \eta \right] \ge 1 - \delta \] 
where $\eta$ is the exact number of solutions for the input formula, 
$\epsilon>0$ is the tolerance and $(1- \delta)$ is the confidence. 
Observe that $\eta$ corresponds exactly to the numerator 
$|\{\mbf{x}\in\mbb{F}:\kappa(\mbf{x})=c\land(\mbf{x}_{\fml{X}}=\mbf{v}_{\fml{X}})\}|$ 
of the conditional probability 
$\prob_{\mbf{x}}(\kappa(\mbf{x})=c\,|\,\mbf{x}_{\fml{X}}=\mbf{v}_{\fml{X}})$ 
in~\cref{eq:wpaxp}, and the bounded approximation $\eta^\ast$ is subsequently 
a PAC approximation of  
$|\{\mbf{x}\in\mbb{F}:\kappa(\mbf{x})=c\land(\mbf{x}_{\fml{X}}=\mbf{v}_{\fml{X}})\}|$.
Additionally, it is plain that  $|\{\mbf{x}\in\mbb{F}:(\mbf{x}_{\fml{X}}=\mbf{v}_{\fml{X}})\}| \ge 1$  
(equals 1 when all features are fixed to values of $\mbf{v}$, i.e.\ $\fml{X} = \fml{F}$), hence 
we have a probability bounded approximation  
$\sfrac{\eta^\ast}{|\{\mbf{x}\in\mbb{F}:\kappa(\mbf{x})=c\land(\mbf{x}_{\fml{X}}=\mbf{v}_{\fml{X}})\}|}$ with 
a tolerance $\epsilon$ and confidence $(1 - \delta)$,
\begin{align*}
	 \frac{p}{1+\epsilon} 
	  \le 
	 \frac{\eta^\ast}{|\{\mbf{x}\in\mbb{F}:\kappa(\mbf{x})=c\land(\mbf{x}_{\fml{X}}=\mbf{v}_{\fml{X}})\}|} 
	     \le  
	 (1+\epsilon)\times p
\end{align*}	 
where $p$ denotes 
$\prob_{\mbf{x}}(\kappa(\mbf{x})=c\,|\,\mbf{x}_{\fml{X}}=\mbf{v}_{\fml{X}})$.

\subsection{Monte-Carlo Sampling}
Our second proposed approach is Monte Carlo sampling (MC-sampling) over 
$\mbf{x}_{\fml{S}}\sim\mbb{F}$ (unifor\-mly sample from $\mbb{F}$ 
restricted to variables of $\fml{F}\setminus\fml{S}$ such that 
$\bigwedge\nolimits_{i\in\fml{S}} x_i=v_i$)   and test $\kappa$ 
on each sample as input --- this will give the 
count  of samples that are in the same class as input 
$\mbf{v}$ (i.e.\ $\kappa(\mbf{x}) = \kappa(\mbf{v})$), namely 
not adversarial examples.
In the same vein as for approximate model counting method, we propose 
an ($\epsilon,\delta$)-approximation solution. 
Subsequently, The number of samples to generate is identified by 
the standard Chernoff bounds~\cite{Mitzenmacher-17} expressed with the parameters 
of tolerance $\epsilon$ and confidence $\delta$.
Concretely,  let $X_i$ be a 0--1 random variable denoting 
the result of the trial with sample $\mbf{z}_i$, where $X_i = 1$ iff 
$\mathbbm{1}_{(\kappa(\mbf{z}_i) = \kappa(\mbf{v}))}(\mbf{z}_i) = 1$ and 
$X_i = 0$ otherwise.
Let $X$ be the random variable denoting the number of trials 
in $X_1, X_2, \dots, X_N$ for which 
$\mathbbm{1}_{(\kappa(\mbf{z}_i) = \kappa(\mbf{v}))}$ returns 1. 
Then, 

\begin{proposition}[Hoeffding~\cite{game-theory-book}]
\label{prop:hoeffding}%
Given independent 0--1 random variables $X_i$,  $X =  \frac{1}{N}\sum\nolimits_{i=1}^{N} X_i$, 
the expected value  $\mu = \mbb{E}(X)$
and  $\epsilon > 0$,  $\prob(\mid X - \mu \mid \ge \epsilon ) \le 2 e^{-2\epsilon^2 N}$.
%
\end{proposition}

Let $\delta > 0$ such that $2 e^{-2\epsilon^2 N} \le \delta$, then we have: 
\[ N \ge \frac{1}{2\epsilon^2} \log (\frac{2}{\delta}) \]

Note that $N$ is the minimum number of samples that needed to be 
(randomly) drawn in order to guarantee the Hoeffding bounds on $\mu$,   
%
\[ 
    \prob\left[ X - \epsilon \le \mu \le X + \epsilon \right]  \ge 1 - \delta 
\]

Clearly, we can estimate the probability of an explanation  through
sampling by applying the Hoeffding bounds to empirically estimate 
the mean $X$ in $N$ trial.
Hence, the estimate $\mu$ represents the ($\epsilon,\delta$)-approximation  
probability of the explanation precision 
$\prob_{\mbf{x}}(\kappa(\mbf{x})=\kappa(\mbf{v})\,|\,\mbf{x}_{\fml{X}}=\mbf{v}_{\fml{X}})$.

Observe that for MC-sampling method, we have 
an additive bound (i.e. $\pm \epsilon$); whilst in approximate 
model counting method, 
we provide a PAC precision of a factor $(1 + \epsilon)$, which is 
a stronger bound, thus a more accurate approximation 
of the true probability of the explanation.
Nevertheless, as shown in the results (see \cref{sec:res}), for small 
$\epsilon$ and $\delta$ that is not close to zero (e.g. $\delta \simeq 0.05$), 
MC-sampling technique shows similar precisions as approximate 
model counting and notably significantly faster to compute.

\begin{algorithm}[ht]
    \providecommand{\Tau}{\mathrm{T}}
\begin{flushleft}
\hspace*{0.2em}
\textbf{Input}: {Parameters: $\fml{E}$, $\Tau$; 
		       Hyperparameters $\epsilon$, $\delta$} \\
\hspace*{0.2em}
\textbf{Output}: {LmPAXp $\fml{S}$ s.t. 
		$\prob\left[ \prob_\mbf{x}(\fml{S}) \ge \Tau - \epsilon \right] \ge 1- \delta$ }
\end{flushleft}
\begin{algorithmic}[1]
  \Procedure{$\delxp$}{$\fml{E},\Tau; \epsilon, \delta$}
  \State{$\fml{S} \gets \fml{F}$}
  \State{$\Phi \gets  \msf{encode}(\fml{E})$} 
  \For{$i \in \fml{F}$ } 
      \State{$mc \gets \msf{approxCount}(Sol(\Phi_{\mid\fml{S}\setminus\set{i}}), \epsilon, \frac{\delta}{m})$}
      \State{$pr \gets \sfrac{mc}{\prod\nolimits_{j\in\fml{F}\setminus(\fml{S}\setminus\{i\})} ||\mbb{D}_j||}$}
      \If{$pr \ge \Tau $}
      \Comment{ $\wpaxp(\mbf{S})$?}
      \State{$\fml{S} \gets \fml{S}\setminus\set{i}$}
      \EndIf
  \EndFor
  \State{\bfseries{return}~{$\fml{S}$}}
\EndProcedure
\end{algorithmic}

    \caption{Deletion-based method for computing LmPAXp}
    \label{alg:del}
\end{algorithm}

\subsection{Computing Locally-Minimal PAXp's}
%
\cref{alg:del} depicts a (deletion-based) linear search method 
 for computing a locally-minimal PAXp.
As shown, to compute one LmPAXp $\fml{S}$, one can starts from
an initial set $\fml{F}=\{1,\ldots,m\}$ containing all features of the data 
or immediately from an AXp $\fml{X}\subset\fml{F}$  and iteratively 
removes least contributing features while it is
safe to do so, i.e.\ while Eq.~\eqref{eq:wpaxp} holds for the 
resulting set.
Concretely, the procedure computes for $\fml{S}$ an approximation 
of the number of solutions that admits the updated subformula 
$\Phi \wedge \bigwedge\nolimits_{j\in\fml{S}} (x_j=v_j)$
(or denoted $\Phi_{\mid\fml{S}}$ representing the classifier restricted to $\fml{S}$) 
and subsequently calculate the probability 
$\prob_{\mbf{x}}(\kappa(\mbf{x})=\kappa(\mbf{x})\,|\,\mbf{x}_{\fml{S}}=\mbf{v}_{\fml{S}})$ 
and checks if it is still greater than $\Tau$. 
The model counting approach and Monte-Carlo sampling method are 
implemented in the routine procedure $\msf{approxCount}$ --- the choice 
for the approach to use is configured in the parameters of the function. 
Moreover,  we implement an heuristic using Monte-Carlo sampling 
approach to order the features from least to highest heuristic score, which allows  
in practice to obtain smaller probabilistic explanations.
Note that for FFAXp explanations, one can apply the FFA scores in the heuristic 
order of the features. 
Observe that features not belonging to an AXp do not contribute 
in the decision of $\kappa(\mbf{v})$  and thus can be safely removed 
at the initialisation step, which allows us to improve the performance 
of \cref{alg:del}, i.e. initialize $\fml{S}$ to some AXp 
$\fml{X}\subseteq\fml{F}$. Our empirical results showcase that 
the fractions between the AXp's size and the resulting locally-minimal AXp's 
are significantly large.  

Note that besides linear search algorithms  for solving function problems 
in propositional logic and constraint programming~\cite{msjm-aij17-XYZ,msjb-cav13}, 
QuickExplain~\cite{junker-aaai04} 
is another alternative approach that could be considered in this problem.

\section{Experiments} \label{sec:res}
This section presents a summary of empirical assessment 
of computing  Probabilistic Abductive explanations  for the case study 
of RF and  BNN classifiers trained on some of the widely studied datasets. 
Moreover, we assess our approach with the baseline method for computing 
minimal probabilistic approach for decision trees (DTs)~\cite{ihincms-ijar23}.
This comparison allows us to validate the accuracy and effectiveness 
of our approximation method with the exact method for scalable family 
of classifiers.  

\sisetup{parse-numbers=false,detect-all,mode=text}
\setlength{\tabcolsep}{3pt}
\let\lpr\undefined
\let\rpr\undefined
\newcommand{\lpr}{(}
\newcommand{\rpr}{)}

\setcounter{table}{3}

\begin{table}[ht]
\centering
\resizebox{0.98\columnwidth}{!}{
  \begin{tabular}{l S[table-format=2]S[table-format=5] 
  S[table-format=1.1]S[table-format=2.2]S[table-format=2.2]
  S[table-format=1.1]S[table-format=2.2]S[table-format=1.2]}
\toprule[1.2pt]
\multirow{2}{*}{\bf Dataset} & \multicolumn{2}{c}{ } & \multicolumn{3}{c}{\bf  MinPAXp } & \multicolumn{3}{c}{\bf  LmPAXp} \\
  \cmidrule[0.8pt](lr{.75em}){4-6}
  \cmidrule[0.8pt](lr{.75em}){7-9}
& {\bf m }  &  {\bf \#S}  &  {\bf Len }  &  {\bf Prec}  &  {\bf  Time}  &   {\bf Len }  &  {\bf Prec}  &  {\bf Time}   \\
\toprule[1.2pt]

adult & 12 & 4848 &  5.6 & 94.6 & 4.64 & 5.7 & 94.79 & 0.17 \\
dermatology & 34 & 292 & 4.0 & 98.8 & 0.35 & 4.0 & 98.86 & 0.13 \\
kr-vs-kp & 36 & 2556 & 3.2 & 95.4 & 0.92 & 3.3 & 95.40 & 0.21 \\
letter & 16 & 14934 & 7.7 & 97.7 & 16.35 & 7.7 & 97.69 & 0.25 \\
soybean & 35 & 498 & 6.1 & 98.1 & 0.94 & 6.1 & 98.21 & 0.22 \\
spambase & 57 & 3368 & 2.4 & 92.4 & 2.15 & 2.6 & 93.12 & 0.93 \\
factors & 216 & 1595 & 5.3 & 98.30 & 1.64 & 5.3 & 98.28 & 1.28 \\
texture & 40 & 4378 & 5.4 & 98.5 & 2.20 & 5.3 & 98.90 & 2.06 \\

\bottomrule[1.2pt]
\end{tabular}
}
\caption{%
  Assessing LmPAXp's for DTs and  compare with {\it minimum} PAXp's 
  \cite{ihincms-ijar23}.  Threshold $\Tau$ was fixed to 0.9.
Columns {\bf m} shows  the number of features in the datasets  and {\bf \#S}  
reports the number tested instance for each dataset.
Columns {\bf Len} and  {\bf Prec}  report, resp., the average length 
and precision/accuracy of computed LmPAXp's (resp.\ MinPAXp) and 
{\bf Time} shows the average runtime for computing 
such explanations.
} 
\label{tab:DTs}
\end{table}

\sisetup{parse-numbers=false,detect-all,mode=text}
\setlength{\tabcolsep}{5pt}
\let\lpr\undefined
\let\rpr\undefined
\newcommand{\lpr}{(}
\newcommand{\rpr}{)}

\begin{table*}[ht]
\centering
\resizebox{0.94\textwidth}{!}{
  \begin{tabular}{l>{\lpr}S[table-format=2.0,table-space-text-pre=\lpr]S[table-format=2.0,table-space-text-post=\rpr]<{\rpr} 
  S[table-format=2.1]S[table-format=2.2] 
  S[table-format=2.1]S[table-format=2]S[table-format=1.2]S[table-format=2.2] || 
  S[table-format=2.1]S[table-format=3.2] 
  S[table-format=2.1]S[table-format=2]S[table-format=1.2]S[table-format=2.2]  
 }
\toprule[1.2pt]
\multirow{2}{*}{\bf Dataset} & \multicolumn{2}{c}{\multirow{2}{*}{\bf (m,~K)}}  & 
\multicolumn{2}{c}{\bf  AXp }  & \multicolumn{4}{c}{\bf  LmPAXp} & 
\multicolumn{2}{c}{\bf  FFAXp }  & \multicolumn{4}{c}{\bf  LmPFFAXp} \\
\cmidrule[0.8pt](lr{.75em}){4-5}
\cmidrule[0.8pt](lr{.75em}){6-9}
\cmidrule[0.8pt](lr{.75em}){10-11}
\cmidrule[0.8pt](lr{.75em}){12-15}
& \multicolumn{2}{c}{}  & {\bf Len }  &  {\bf Time}  &  {\bf Len }  & {\bf \% }  & {\bf Prec }  &  {\bf Time } &
 {\bf Len }  &  {\bf Time}  &  {\bf Len }  & {\bf \% }  & {\bf Prec }  &  {\bf Time }  \\
\toprule[1.2pt]

adult  &  12  &  2  &  5.6  &  0.19  &  3.2  &  57  &  0.97  &  0.99 &  6.0  &  0.04  &  5.0  &  83  &  0.99  &  0.67 \\
agaricus  &  22  &  2  &  10.0  &  0.06  &  3.3  &  33  &  0.96  &  3.78 &  20.0  &  24.66  &  3.4  &  17  &  0.97  &  16.81\\
compas  &  11  &  2  &  5.6  &  0.12  &  3.6  &  64  &  0.97  &  0.77 &  10.0  &  0.37  &  3.9  &  39  &  0.97  &  2.75 \\
german  &  21  &  2  &  12.3  &  0.36  &  3.2  &  26  &  0.97  &  5.40 &  21.0  &  155.64  &  3.4  &  16  &  0.97  &  16.01 \\
heart-c  &  13  &  2  &  5.5  &  0.07  &  3.0  &  55  &  0.97  &  0.39 &  13.0  &  1.99  &  2.9  &  22  &  0.97  &  2.13 \\
ionosphere  &  34  &  2  &  21.5  &  0.06  &  2.9  &  13  &  0.96  &  6.59 &  33.0  &  13.17  &  2.9  &  9  &  0.97  &  14.68 \\
kr\_vs\_kp  &  36  &  2  &  8.1  &  0.08  &  3.5  &  43  &  0.98  &  0.84 &  33.0  &  3.58  &  3.3  &  10  &  0.99  &  15.46 \\
lending  &  9  &  2  &  2.2  &  0.10  &  1.7  &  77  &  0.99  &  0.15 &  2.0  &  0.03  &  1.7  &  85  &  0.99  &  0.12 \\
mofn\_3\_7\_10  &  10  &  2  &  3.1  &  0.06  &  2.8  &  90  &  0.99  &  0.13 &  5.0  &  0.01  &  3.4  &  68  &  0.99  &  0.29 \\
mushroom  &  22  &  2  &  9.2  &  0.07  &  4.2  &  46  &  0.97  &  2.30 &  20.0  &  4.72  &  4.3  &  22  &  0.97  &  11.16 \\
parity5+5  &  10  &  2  &  6.8  &  0.23  &  6.4  &  94  &  0.99  &  0.38 &  9.0  &  0.03  &  6.7  &  74  &  0.99  &  0.82 \\
recidivism  &  15  &  2  &  7.9  &  0.48  &  5.8  &  73  &  0.97  &  1.75 &  15.0  &  0.43  &  5.7  &  38  &  0.97  &  8.49 \\
segmentation  &  19  &  7  &  8.1  &  0.39  &  4.9  &  60  &  0.98  &  0.76 &  17.0  &  10.42  &  4.9  &  29  &  0.98  &  4.11 \\
soybean  &  35  &  18  &  15.6  &  2.05  &  8.9  &  57  &  0.97  &  4.07 &  34.0  &  230.38  &  8.7  &  26  &  0.97  &  24.36 \\
tic\_tac\_toe  &  9  &  2  &  5.0  &  0.09  &  3.3  &  66  &  0.97  &  0.58 &  9.0  &  0.03  &  3.4  &  38  &  0.97  &  2.17 \\
twonorm  &  20  &  2  &  10.4  &  0.05  &  4.4  &  42  &  0.97  &  1.35 &  15.0  &  0.04  &  8.1  &  54  &  0.98  &  1.97 \\
vote  &  16  &  2  &  5.6  &  0.07  &  2.6  &  46  &  0.98  &  1.16 &  16.0  &  1.26  &  2.5  &  16  &  0.98  &  10.65 \\

\bottomrule[1.2pt]
\end{tabular}
}
\caption{%
  Detailed performance evaluation of computing LmPAXp and 
  LmPFFAXp explanations for RFs.
The table shows results for 17 datasets containing binary and 
categorical data. The maximum tolerance  was fixed to 0.05, i.e.\ 
$\Tau = 0.95$.
%
(Remaining columns keep the same meaning as in~\cref{tab:DTs}.)
} 
\label{tab:RFs-sat}
\end{table*}

\paragraph{Experimental setup.}
The experiments are conducted on Intel Core~i5-10500 3.1GHz CPU 
with 16GByte RAM running Ubuntu 22.04. LTS.
A time limit for each single call to the approximate counter is fixed 
to $120$ seconds and a general time limit for delivering an explanation 
is set to $600$ seconds; whilst the memory limit is set to 4 GByte. 

\paragraph{Benchmarks.}
The assessment of RFs is performed on a selection of 17 publicly available 
datasets, which originate from UCI Machine Learning Repository
\cite{uci} and Penn Machine Learning Benchmarks~\cite{Olson2017PMLB}.
Benchmarks comprise  binary and multidimensional classification datasets 
and include binary or/and categorical datasets. 
(Categorical features are encoded into bit-vectors and handled as 
a group of attributes representing their original features when computing 
the explanations.)
%
%
When training RF classifiers for the selected datasets, we used 80\% of
the dataset instances (20\% used for test data). For assessing 
explanation tools, we randomly picked fractions of the dataset,
depending on the dataset size. 
%
%
Besides, the assessment of BNNs is performed on image datasets 
of MNIST digits~\cite{LeCun98}. 
First, we binarized the data (convert pixel values into 0 or 1) and 
generated  7 binary class datasets from the original multi-class 
(from 0 to 9) dataset. Namely, each resulting binary dataset contains 
images of 2 labels $c_1$ \emph{versus} $c_2$ 
(e.g. mnist-0vs8 comprises samples of class 0 and 8).
We considered different tunings of the training parameters 
so that we obtain the best training and test accuracy of each 
benchmark. 
The precision threshold $\Tau$ of LmPAXp is fixed to 95\% for 
all considered benchmarks in RFs and 99\% for BNN assessment.
As results, to guarantee high probabilistic precision of the explanations  
we implement the approximate model counting metric for BNNs; whilst 
we apply Monte-Carlo sampling metric for RFs to improve the effectiveness 
and scalability of our solution. 
Note that, our observation on preliminary results demonstrate that for 
$\Tau \leq 0.97$ both metrics (sampling and model counting) return  
similar results.
Lastly, for assessment on DTs we re-used the exact benchmarks (i.e.\ datasets
and DT models) as in~\cite{ihincms-ijar23}  that were  provided by the authors.

\paragraph{Prototypes implementation.}
We developed a reasoner for RFs as a Python script. The script 
implements the SAT-based approach for explaining RFs proposed 
by~\cite{ims-ijcai21}, in addition the deletion/progression based  
algorithms described above for computing LmPXAp. 
Furthermore, PySAT~\cite{imms-sat18} is used to instrument incremental 
SAT oracle calls, and the oracles 
ApproxMC\footnote{\url{https://github.com/meelgroup/approxmc}} 
\cite{ChakrabortyMV16,SoosM19,SoosGM20,MeelA20} and 
ApproxMCPB\footnote{\url{https://github.com/meelgroup/approxmcpb}}  
\cite{YangM21,YangM23} to instrument (approximate) model counting calls
on SAT and pseudo-Boolean formulas.
Besides, we reused the BNN encoding of~\cite{rinard-nips20}\footnote{
\url{https://github.com/jia-kai/eevbnn}} to 
implement both SAT and pseudo-Boolean -based encodings for our 
formulation of computing explanations.  
Similarly to RF encoding, we used PySAT to generate BNN SAT-based encoding, 
and additionally  we used the Python package PyPBLib~\cite{pypblib} of 
PBLib~\cite{pblib.sat2015}, 
which is integrated in PySAT,  to generate the pseudo-Boolean encoding.
(Note that PyPBLib/PBLib  by design provides efficient clausal form encodings 
of pseudo-Boolean formulas. Accordingly, we used it on purpose to generate 
the pseudo-Boolean constraints and convert them into CNF formulas when 
SAT encoding is selected.) 
Moreover, Minisat (or Glugose3) SAT solver is instrumented for the resolution  
of the CNF formulation, whilst pseudo-Boolean oracle Roundingsat\footnote{%
Roundingsat is a pseudo-Boolean solver augmented with 
the LP solver SoPlex~\cite{soplex}.}~\cite{Nordstrom-ijcai18}  
is instrumented to solve the pseudo-Boolean encodings.
We underline that our observations on preliminary assessments, and 
also as pointed out in~\cite{rinard-nips20}, pseudo-Boolean encoding 
and Roundingsat oracle shows (slightly) better performances than 
SAT oracles. 
As a result, we solely report the experimental results of the pseudo-Boolean 
encodings of the selected BNN benchmarks.    
%

\paragraph{Baseline.}
In order to validate our approach in terms of explanation accuracy and size, 
we used as a baseline the dedicated approach for computing MinPAXp's 
\cite{ihincms-ijar23} for DTs. 
As can be seen from \autoref{tab:DTs}, our method based on MC sampling shows (almost)
comparable results with~\cite{ihincms-ijar23}, in terms of explanation length and precision. 
Furthermore, although  the computation of MinPAXp's shows better runtimes, we observe that 
LmPAXp's  runtimes  are also very small (i.e. less than 1s (resp.\ 2s)  
for 6/8 (2/8)). 
Hence, one can argue that our {\it general} solution is competitive with model specific 
techniques and these additional results empirically validate  explanations 
faithfulness delivered by our method.

\sisetup{parse-numbers=false,detect-all,mode=text}
\setlength{\tabcolsep}{5pt}
\let\lpr\undefined
\let\rpr\undefined
\newcommand{\lpr}{(}
\newcommand{\rpr}{)}

\begin{table*}[ht]
\centering
\resizebox{0.85\textwidth}{!}{
  \begin{tabular}{l S[table-format=2.1]S[table-format=1.2] 
  S[table-format=2.1]S[table-format=2]S[table-format=3.2]S[table-format=2]
  S[table-format=2.1]S[table-format=1.2] 
  S[table-format=2.1]S[table-format=2]S[table-format=3.2]S[table-format=2] }
\toprule[1.2pt]
\multirow{2}{*}{\bf Dataset} & \multicolumn{2}{c}{\bf  AXp } & \multicolumn{4}{c}{\bf  LmPAXp } 
& \multicolumn{2}{c}{\bf  FFAXp } & \multicolumn{4}{c}{\bf  LmFFAXp } \\
  \cmidrule[0.8pt](lr{.75em}){2-3}
  \cmidrule[0.8pt](lr{.75em}){4-7}
  \cmidrule[0.8pt](lr{.75em}){8-9}
  \cmidrule[0.8pt](lr{.75em}){10-13}
& {\bf Len }  &  {\bf Time}  &  {\bf Len } & {\bf \%}  &  {\bf Time}  &  {\bf \#TO }  
& {\bf Len }  &  {\bf Time}  &  {\bf Len } & {\bf \%}  &  {\bf Time}  &  {\bf \#TO }    \\
\toprule[1.2pt]

%
mnist-0vs8  &  28.8  &  0.51  &  10.1  &  35  &  345.24  &  11 &  42.68  &  1.40  &  10.2  &  24  &  443.95  &  12 \\
mnist-1vs6  &  25.6  &  0.52  &  9.0  &  35  &  125.69  &  0  &  44.24  &  1.79  &  9.0  &  20  &  281.40  &  0 \\
mnist-2vs6  &  18.6  &  0.47  &  12.0  &  65  &  27.76  &  0 &  27.62  &  0.18  &  12.2  &  44  &  75.50  &  0  \\
mnist-1vs8  &  29.3  &  0.50  &  12.5  &  43  &  150.10  &  0  &  43.78  &  1.51  &  12.7  &  29  &  279.08  &  1  \\
mnist-2vs8  &  29.6  &  0.48  &  12.6  &  43  &  169.04  &  0  & 47.66  &  1.57  &  15.2  &  26  &  312.76  &  0 \\
mnist-3vs8  &  29.7  &  0.48  &  11.8  &  40  &  118.16  &  0 &  43.14  &  0.78  &  13.3  &  31  &  249.88  &  0 \\
mnist-4vs9  &  26.1  &  0.47  &  9.8  &  38  &  87.81  &  0  &  38.32  &  0.67  &  9.7  &  25  &  182.84  &  0 \\

\bottomrule[1.2pt]
\end{tabular}
}
\caption{%
  Detailed performance evaluation of computing locally-minimal PAXp for BNNs 
  trained on (binarized) MNIST datasets.
%
%
The  maximum tolerance is fixed to 0.01, i.e. $\Tau = 0.99$.
%
%
Column {\bf \#TO} reports the number timeouts reported over 100 tests 
for the deletion-based algorithm.
(Remaining columns keep the same meaning as in~\cref{tab:DTs}.)
} 
\label{tab:BNNs-pb}
\bigskip
\end{table*}

\paragraph{MC-Sampling.}
\autoref{tab:RFs-sat}  summarizes the results of assessing 
the probabilistic explanation succinctness of RFs using MC-Sampling 
technique. 
Notably, results shown \autoref{tab:RFs-sat} reports the average lengths 
of LmPAXp's {\it versus} with AXp's, and FFAXp's {\it versus} LmFFAXp's  
computed with \cref{alg:del} using MC-sampling.
As can be observed from \autoref{tab:RFs-sat}, with a few exceptions 
(2 binary datasets out of 17, i.e. {\it mofn\_3\_7\_10} and {\it parity5+5})
$\lmpaxp$ provides shorter
explanations for all tested datasets --- this is illustrated for example, 
in {\it ionosphere} and {\it german} datasets 
the average size of probabilistic abductive explanations is, resp., 86.6\% and
73.9\% smaller than plain abductive explanations, with an average precision 
of 0.97 on all tested samples.
Moreover, for the majority of the datasets the average precision is 0.97 or 
higher, in total 16 out 17 datasets (i.e. 8 out 17, 5 out 17 and 3 out, resp.\ 
show an average precision of 0.97, 0.98 and 0.99) and 1 dataset 
has an average precision of 0.96.
Besides, according to our expectations, the average runtimes of our 
sampling-based approach are small --- the minimum 
average runtime reported is 0.13 seconds, while the maximum does not 
exceed 4.8 seconds.    
Unsurprisingly FFAXp's are even larger that AXp's, however we observe 
that LmPFFAXp average length's are (almost) similar to LmPAXp's. 
As a result, one can remark a significant gain of size for  LmPFFAXp's, 
e.g.\ for {\it agaricus}  the average length of FFAXp is 20 (two time the average 
length of AXp's) and the average length of LmPFFAXp's is 3.4 
(tightly close to the average size of LmPAXp's, which is 3.3), i.e. a gap 
of 83\%; and for {\it soybean}  we have LmPFFAXp average size that 
represents only 26\% of the initial FFAXp average size.

\paragraph{ApproxMCounting.}
Recall that for high accuracy of probabilistic explanations, 
ApproxMC is more faithful offering higher probabilistic guarantees 
than MC-Sampling.
We evaluate our ApproxMC technique for computing LmPAXp's/LmPFFAXp's 
on more challenging family of classifiers than MC-Sampling assessment.  
The aim is to showcase that for classifier like BNNs trained on image 
data, we are able to obtain more succinct formal probabilistic explanations 
than formal explanations, with high probability of precision. 
The counterpart is the runtime of ApproxMC that is usually higher than 
MC-Sampling, nevertheless our results show it is possible to reduce 
significantly the size of AXp's in a reasonable time. 

The obtained results on BNNs, summarized in \autoref{tab:BNNs-pb}, 
show significant size reduction in the explanations of $\lmpaxp$ 
compared to $\axp$ while 
offering a very high precision  $\Tau = 0.99$.  
We observe a difference of average length of 34.5\% up to 68.2\% between 
LmPAXp's and AXp's, e.g. the average number of pixels for $\axp$ in 
{\it mnist-0vs8}  is ~29, whereas for $\lmpaxp$ it is ~10 pixels, which 
represents 33.6\% of its average AXp size.
Similarly as RFs, we observe a large gain of size between FFAXp's and 
LmFFAp's. 
In fact, FFAXp's by their nature are larger than AXp's, 
in contrast we observe that the resulting LmFFAXp's are always 
as succinct as LmPAXp's, even though their feature are not 
identical/similar, since the initial sets of features to inspect are 
different and the order for analyzing the features are also different 
(i.e. for LmPFFAXp's we apply the $\ffa$ and for AXp's either lexicographic 
or heuristic score computed with sampling).    
In terms of runtimes, the averages reported for LmPAXp's vary 
from  27.76 seconds to 345.24 seconds (avg.\ 146.25 seconds), with 
6 out of 7 BNNs that successfully returned an explanation without exceeding 
the timeout fixed for  ApproxMC oracle, whilst {\it mnist-0vs8} reports 11 out 
of 100 test that have failed to terminate.
For LmPFFAXp's, we remark that runtimes are slightly higher than LmPAXp's 
which is not surprising since that the algorithm performs more iterations  
for larger initial set of features to inspect (i.e. FFAXp's are larger than AXp's).
Also, we observe that out of 7 BNNs 5 show average runtimes less than 
300 seconds (5 minutes) and a total average of 260.77 seconds. 
%

\section{Conclusions} \label{sec:conc}

Formal explainability witnessed significant progress in recent years.
However, some challenges remain, including the size of explanations.
Probabilistic abductive explanations (PAXps) represent a solution to
curb explanation size. Unfortunately, their exact computation is
impractical.
An alternative to PAXps are locally minimal probabilistic abductive
explanations (LmPAXps), which recent work showed, for some families of
classifiers, that LmPAXps match in practice PAXps.
Motivated by these results, this paper proposes two algorithms for
approximating the computation of LmPAXps for complex ML models.
The experimental results reveal two possible use-cases, one for each
of the proposed algorithms.
In application domains where probabilistic formal explanations suffice, i.e.\ 90\%
to 97\% precision, the paper proposes a novel algorithm based on
MC-sampling, which scales to \emph{any} classifier.
Alternatively, in domains where explanations with high accuracy
(i.e.\ around 99\%) are of interest,  the paper proposes another
novel algorithm based on approximate model counting. 
Future work will consider the use of approximate model counting for
families of classifiers beyond boolean domains. This will require
integration of approximate model counting with SMT reasoners.

\newtoggle{mkbbl}

\settoggle{mkbbl}{false}

\iftoggle{mkbbl}{
  \bibliographystyle{abbrv}
  \bibliography{team,refs}
}{

}

\end{document}